\documentclass[lettersize,journal]{IEEEtran}
\usepackage{amsmath,amsfonts,nccmath}
\usepackage{algorithmic}
\usepackage{algorithm}
\usepackage{array}
\usepackage[caption=false,font=normalsize,labelfont=sf,textfont=sf]{subfig}
\usepackage{textcomp}
\usepackage{stfloats}
\usepackage{url}
\usepackage{verbatim}
\usepackage{graphicx}
\usepackage{cite}
\usepackage{pdfpages}
\usepackage{enumitem}

\usepackage{amssymb}
\usepackage{amsthm}
\usepackage{tikz}
\usepackage{forest}
\usetikzlibrary{trees, positioning,shapes,shadows,arrows.meta}
\usetikzlibrary{mindmap,shadows,backgrounds}
\usepackage{hyperref}
\usepackage{booktabs}
\usepackage{multirow}
\usepackage{titlesec}
\newtheorem{definition}{Definition}
\titlespacing*
    {\subsection}
    {0pt}
    {1ex}
    {1ex}

\begin{document}

\title{Exploring Incremental Unlearning: Techniques, Challenges, and Future Directions}


\author{Sadia Qureshi, Thanveer Shaik,  Xiaohui Tao, Haoran Xie, Lin Li, Jianming Yong, and Xiaohua Jia
\thanks{Sadia Qureshi, Thanveer Shaik and Xiaohui Tao are with 
the School of Mathematics, Physics and Computing, University of Southern Queensland, Australia (e-mail: Sadia.Qureshi@unisq.edu.au, Thanveer.Shaik@unisq.edu.au, Xiaohui.Tao@unisq.edu.au).}
\thanks{Haoran Xie is with the Department of Computing and Decision Sciences, Lingnan University, Hong Kong (e-mail: hrxie@ln.edu.hk)}
\thanks{Lin Li is with the School of Computer Science and Artificial Intelligence, Wuhan University of Technology, China (e-mail: cathylilin@whut.edu.cn)}
\thanks{Jianming Yong is with 
the School of Business, University of Southern Queensland, Australia (e-mail: Jianming.Yong@unisq.edu.au).}
\thanks{Xiaohua Jia is Head \& Chair Professor of CS, City University of Hong Kong, Hong Kong (e-mail: csjia@cityu.edu.hk).}
}

\markboth{Journal of \LaTeX\ Class Files,~Vol.~14, No.~8, August~2021}%
{Shell \MakeLowercase{\textit{et al.}}: A Sample Article Using IEEEtran.cls for IEEE Journals}


\maketitle

\begin{abstract}
The growing demand for data privacy in Machine Learning (ML) applications has seen Machine Unlearning (MU) emerge as a critical area of research. As the `right to be forgotten' becomes regulated globally, it is increasingly important to develop mechanisms that delete user data from AI systems while maintaining performance and scalability of these systems. Incremental Unlearning (IU) is a promising MU solution to address the challenges of efficiently removing specific data from ML models without the need for expensive and time-consuming full retraining. This paper presents the various techniques and approaches to IU. It explores the challenges faced in designing and implementing IU mechanisms. Datasets and metrics for evaluating the performance of unlearning techniques are discussed as well. Finally, potential solutions to the IU challenges alongside future research directions are offered. This survey provides valuable insights for researchers and practitioners seeking to understand the current landscape of IU and its potential for enhancing privacy-preserving intelligent systems.

\end{abstract}


\section{Introduction}

Machine Unlearning (MU) is an evolving concept that aims to address several critical challenges in the domain of Machine Learning (ML) and Artificial Intelligence (AI), particularly with respect to privacy, ethics, and regulatory compliance \cite{kurmanji2024machine}. It focuses on the need to remove or "forget" specific data points that may have been used in training a model, particularly when that data is sensitive or no longer required\cite{doughan2024machine,schelter2021towards}.
The use of personal or sensitive data to train ML models raises significant privacy concerns, especially in  light of rigorous regulations such as the General Data Protection Regulation (GDPR) \cite{zaeem2020effect} and California Consumer Privacy Act (CCPA)\cite{yanamala2023secure}. These regulations often require that individuals have the right to be "forgotten", meaning that their personal data must be removed from systems upon request\cite{bakare2024data}. 

\begin{figure}

    \centering
    \includegraphics[width=1\linewidth]
    {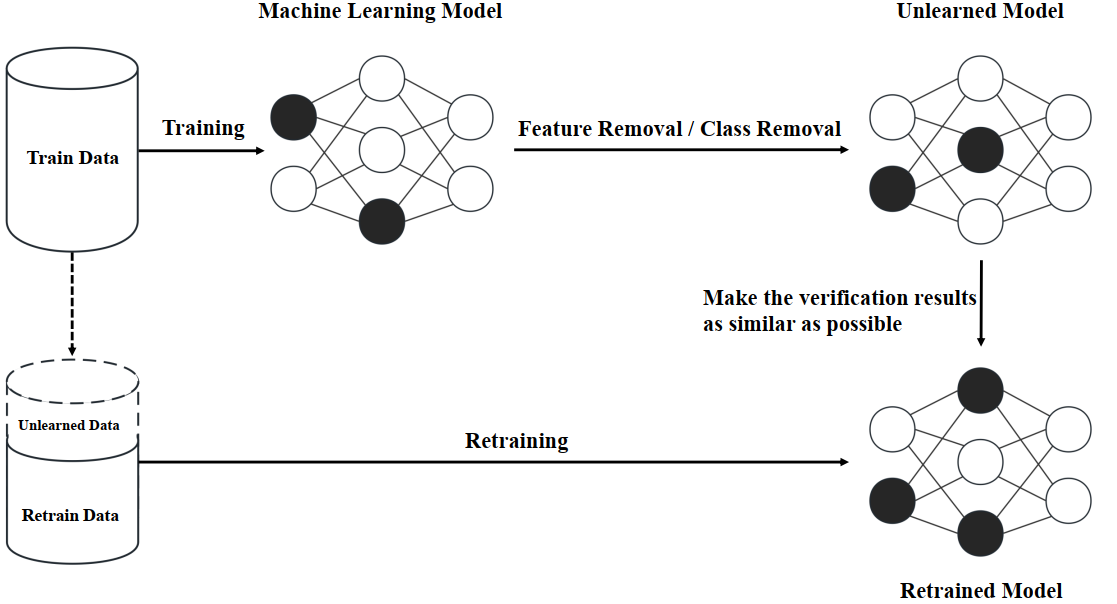}
    \caption{Machine Unlearning Overflow \cite{li2024overview}}
    \label{fig:mu_overflow}
    
\end{figure}

MU offers a solution through the removal of specific data from both the storage archives and the trained models themselves\cite{chundawat2023zero}. This is crucial because deleting data from storage does not automatically ensure that it is no longer present in a trained model. MU techniques attempt to unlearn the influence of this data from the model's parameters, ensuring that both privacy of the model is maintained and that the model complies with data privacy regulations\cite{warnecke2021machine}.
The overall workflow of MU is summarized in Figure \ref{fig:mu_overflow}.

As privacy concerns increase and new regulations are introduced, it is important for ML models to adapt certain changes in the environment. As new data is input, older or less relevant data can be unlearned to keep models up-to-date and relevant without having to retrain them from scratch\cite{jaman2024machine}. For example, a model trained on certain datasets might eventually become non-compliant with new data privacy laws or standards. To resolve these issues, we introduce Incremental Unlearning (IU), which enhances the flexibility of models by allowing dynamic changes to the dataset. IU is a branch of machine unlearning where models can progressively forget data or knowledge without retraining from scratch \cite{weng2024proof}. This capability is especially useful where data privacy, regulatory compliance, and knowledge currency are crucial. This capability could future-proof AI systems against the rapid pace of regulatory changes\cite{hine2024supporting}.

This survey categorizes IU techniques according to the research challenges and goals within the unlearning process. In doing so, we review the differences, relationships and the respective advantages and disadvantages of these techniques. We begin by summarizing IU into four main areas: Reinforcement Unlearning, Continual Learning via Selective Forgetting, Corrective Unlearning, and Federated Unlearning. Since many studies focus on Reinforcement Unlearning, we allot two sections to introduce the corresponding
studies in this area. We classify these unlearning methods into: decremental RL method, environment poisoning method, and quantized reward konditioning (Quark) theory. Next, we discuss continual learning via selective forgetting through selective pruning, decremental matching moment using Gaussian distribution, and selective amnesia. Furthermore, we discuss corrective unlearning though selective synaptic dampening, cognac theory and corrective unranking distillation(CuRD) algorithm. In addition, we introduce the federated unlearning method, which is gaining significant importance in unlearning research. Federated unlearning can be further classified into FedEraser, FedAU and quantized federated learning (Q-FL). 

Following the application of an IU technique, it is critical to verify the unlearning effect. Verification has attracted significant research attention, so we perform a timely review on related research on IU verification. Finally, we consider the essential privacy and security issues in IU by reviewing related state-of-the-art research directed toward privacy and security threats, defenses, and unlearning applications in this space. This literature survey is guided by the following research questions:
\begin{itemize}
    
    \item What are the existing techniques and approaches of IU?
    \item What are the challenges in IU and their potential solutions?
    \item What are the key datasets and evaluation metrics for IU?
    \end{itemize}

At present there are limited literature surveys available on IU, as it is a relatively new research domain focusing. The need to incrementally unlearn scenarios is a growing demand, thus our  survey outlines, examines, and classifies IU methods for researchers . Our objective is to provide the key features and advantages of IU, before introducing a new taxonomy to categorize the latest research in this area. The overview in this paper provides a knowledge foundation for researchers to encourage future innovations and expanding research perspectives. The main contributions of this survey are listed below:
\begin{itemize}
    \item  Our survey aims to provide a comprehensive taxonomy of IU techniques, with a focus on their applicability, efficiency, and scalability.
    
    \item Our paper identifies both key challenges in IU and the potential solutions.
    
    \item Our paper suggests future directions that will enhance the practical deployment of IU.
\end{itemize}

The remainder of the paper is organised as follows. Section \ref{sec:overview} provides key definitions of IU. Section \ref{sec:techniques} covers the techniques and approaches in IU. Section \ref{evaluation and datasets} details the evaluation metrics and datasets, while Section \ref{challenges} discusses the challenges associated with the suggested techniques. In Section \ref{sec:potential and future}, we discuss potential solutions and future directions. Finally, Section \ref{conclusion} concludes the paper.

\section{Overview } 
\label{sec:overview}
Incremental Learning, also known as continual learning, refers to the process in which a ML model is trained on data sequentially with one batch of data or instance at a time rather than the entire dataset at once\cite{gallardo2021self}. This approach is particularly useful in situations where the data is large, or when data is continually generated and must be processed in real time\cite{wu2019large}. Incremental Learning allows a model to learn continuously over time, thus making it more applicable in dynamic environments where the data distribution may change over time such as real-time applications or adaptive systems\cite{wang2024comprehensive}. Mathematically, Incremental Learning can be defined as follows:

\begin{definition}
Let \( \mathcal{D} = \{(x_i, y_i)\}_{i=1}^{N} \) represent a dataset of \( N \) labeled examples, where \( x_i \) is the input and \( y_i \) is the corresponding label.

\textbf{Initial Training:}  
A model \( M_t \) at time \( t \) is trained using an initial dataset \( \mathcal{D}_{\text{init}} \subset \mathcal{D} \). The objective function is:
\[
\theta_t = \arg \min_{\theta} \mathcal{L}(\theta, \mathcal{D}_{\text{init}}),
\]
where \( \mathcal{L} \) is the loss function and \( \theta \) represents the model parameters.

\textbf{Incremental Update:}  
When new data \( \mathcal{D}_{\text{new}} \subset \mathcal{D} \) arrives, the model updates its parameters \( \theta_{t+1} \) while retaining knowledge from \( \mathcal{D}_{\text{init}} \):
\[
\theta_{t+1} = \arg \min_{\theta} \Big( \mathcal{L}(\theta, \mathcal{D}_{\text{new}}) + \lambda \cdot \mathcal{L}_{\text{reg}}(\theta, \mathcal{D}_{\text{init}}) \Big),
\]
where \( \mathcal{L}_{\text{reg}} \) is a regularization term to prevent forgetting, and \( \lambda \) controls the trade-off between learning new knowledge and retaining old knowledge.
\end{definition}

Incremental Learning is a powerful approach in scenarios where data arrives continuously and models must adapt over time, as continuous learning elicits  memory and computational constraints \cite{luo2020appraisal}. Due to its adaptive nature, the approach enables real-time model updates and better supports dynamic data environments. The principle of incrementally learning means that the model doesn't need to forget old information when it learns new information. Instead, knowledge is accumulated as the model is exposed to new data over time without catastrophic forgetting\cite{van2024continual}.

Class Incremental Learning\cite{lan2023elephant,zhou2024class} methods use knowledge distillation\cite{cui2022uncertainty} and exemplar-based approaches\cite{cha2021ssul} to help models retain knowledge from previous classes while learning new classes as well. These methods use strategies to address the issue of catastrophic forgetting that ensure previously learned information is not lost when new data is introduced.
Knowledge distillation helps secure previous knowledge by training new models to present the behaviour of previous models\cite{kang2022class}.
Exemplar-based methods involve retaining a small representative subset of previous data, which helps the model maintain performance on previous classes\cite{he2020incremental}.
By combining these techniques, Incremental Learning can become more effective in real-world applications by enabling models to learn continuously without forgetting important prior knowledge. In a typical learning model, once a model is trained, it becomes difficult to remove specific data points because most models learn patterns across the entire dataset. The solution is Incremental Unlearning (IU), which can efficiently unlearn the influence of specific data points while minimizing the impact on the rest of the model's knowledge. IU involves integrating processes that allow a model to both acquire new knowledge and to discard outdated or irrelevant information. 

The need for IU arises from privacy concerns resulting from regulations like GDPR and CCPA, in which users have the right to be forgotten. If a user requests that their personal data be removed, the system must comply with their request by eliminating the influence of that data on the model's behaviour and predictions. However, it is difficult to retrain the model from scratch, and it can be computationally expensive and time-consuming, especially for large-scale models.

IU focuses on how efficiently a model can unlearn and relearn automatically by selectively forgetting data points and adapting to new data, while maintaining the integrity and reliability of the model. Mathematically, IU is defined as follows:

\begin{definition}
Let \( \mathcal{D} = \{(x_i, y_i)\}_{i=1}^{N} \) represent a dataset of \( N \) labeled examples, where \( x_i \) is the input and \( y_i \) is the corresponding label.

\textbf{Initial Training:}  
Let \( \mathcal{D}_{\text{remove}} \subset \mathcal{D} \) denote the subset of data to be unlearned.

The model \( M_t \) is trained on \( \mathcal{D} \) with parameters \( \theta_t \):
\[
\theta_t = \arg \min_{\theta} \mathcal{L}(\theta, \mathcal{D}).
\]

\textbf{Unlearning Objective:}  
To remove the influence of \( \mathcal{D}_{\text{remove}} \), the model updates its parameters to \( \theta_{t+1} \) while retaining knowledge from \( \mathcal{D} \setminus \mathcal{D}_{\text{remove}} \):
\[
\theta_{t+1} = \arg \min_{\theta} \Big( \mathcal{L}(\theta, \mathcal{D} \setminus \mathcal{D}_{\text{remove}}) + \gamma \cdot \mathcal{L}_{\text{remove}}(\theta, \mathcal{D}_{\text{remove}}) \Big),
\]
where \( \mathcal{L}_{\text{remove}} \) penalizes retention of knowledge from \( \mathcal{D}_{\text{remove}} \), and \( \gamma \) determines the importance of unlearning.
\end{definition}

IU allows models to unlearn data or features incrementally as new data is introduced. Due to ongoing developments, data and environments can change. A model needs to adapt to new information without forgetting previous valuable knowledge.

\subsection{Key features of Incremental Unlearning}\label{sec:Key Aspects of Incremental Unlearning}
There are several key features that clarify the concept behind IU models:

\begin{itemize}
\item Selective Forgetting: The model is capable of forgetting specific data points or patterns that are no longer relevant, without losing the ability to retain useful information\cite{shibata2021learning}.

\item Adaptability: IU enables models to adapt to new data by gradually removing outdated information. This process helps maintain the model's relevance in dynamic environments\cite{gligorea2023adaptive}.

\item Data Privacy and Security: In some use cases (such as medical or bank data), IU can be used to comply with data protection regulations (like GDPR), allowing for deletion of personal data while still retaining the knowledge learned from that data\cite{bae2018security}.

\item Efficient Model: Retraining the model from scratch can be resource-intensive and time-consuming. IU offers a computationally efficient way to update models.

\item Reduced Risk of Model Staleness: With ongoing changes in real-world applications, models can become overfitted to old data, leading to poor decision-making in new situations. IU can avoid this by clearing outdated knowledge, thus maintaining a model's effectiveness.

\end{itemize}

\section{Techniques and Approaches}
\label{sec:techniques}
\begin{figure}
    \centering
    \includegraphics[width=1\linewidth]{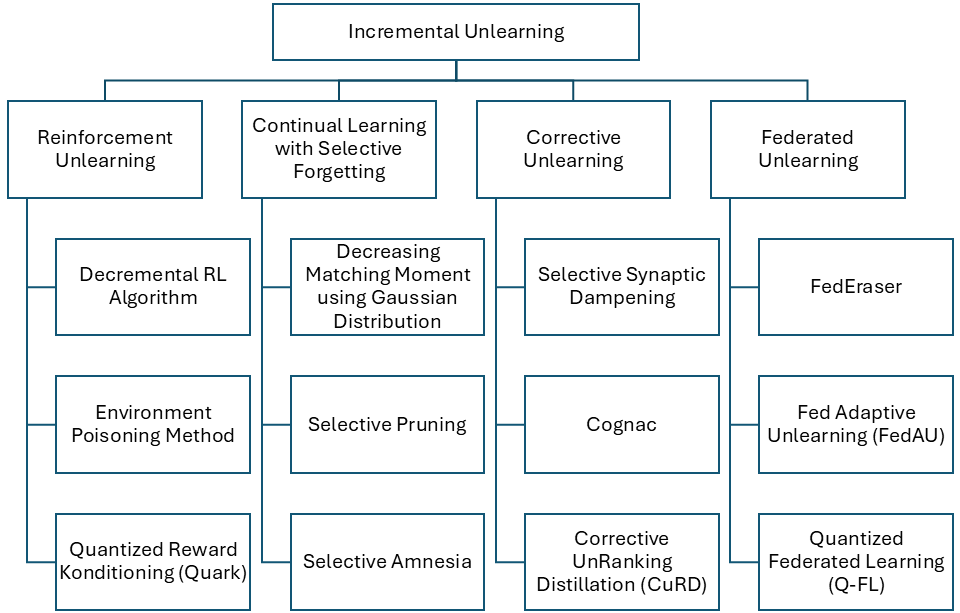}
    \caption{Incremental Unlearning Taxonomy}
    \label{fig:Incremental Unlearning Taxonomy}
\end{figure}

In this section we cover several techniques and approaches adopted by different researchers on MU. We will further discuss how a model incrementally makes runtime decisions in performing certain tasks while preserving its privacy. These techniques are summarized in figure \ref{fig:Incremental Unlearning Taxonomy}.

\subsection{Reinforcement Unlearning}
\label{sec:Reinforcement}
Reinforcement Unlearning (RU) focuses on selectively removing specific learned knowledge from an agent's memory, ensuring that sensitive or outdated information is not retained. In Reinforcement Learning\cite{ladosz2022exploration}, an agent learns to make optimal decisions within an environment to maximize cumulative rewards \cite{samsami2020distributed}. Integrating RU can ensure that the agent forgets definite information, which can be particularly useful in situations where there are privacy concerns or when the learned information becomes irrelevant. 

However, there are several problems associated with RU\cite{ye2024reinforcement}. A significant difficulty lies in linking an environment\cite{martignoni2021did} that is required to be unlearned to its corresponding experience data. This is due to the environment owner's lack of access to the agent-managed experience data\cite{dong2022simple}. Another issue is that excessive unlearning can lead to a decline in the agent's overall performance. Moreover, it is challenging to find a way to evaluate RU effectively. Since the environment owner cannot specify what samples should be unlearned, it makes traditional evaluation methods like membership inference ineffective \cite{dealcala2024comprehensive} \cite{salem2018ml}. In addition, other challenges associated with Reinforcement Learning \cite{dulac2019challenges, ding2020challenges, vinyals2017starcraft} may also be inherently faced when using RU like adaptability and stability in dynamic environment. 

RU can be approached in a number of ways. 
Decremental Reinforcement Learning (DRL) is one such approach that focuses on adapting and updating the learning process by gradually diminishing the influence of older experiences while emphasizing more recent and relevant ones \cite{li2017deep}. The DRL method works in two steps. In the first step, the agent explores its unlearning environment by collecting different samples using a random policy. The second step involves fine-tuning the samples collected by the user. This approach helps the agent build a comprehensive understanding of the environment, which is crucial for effective unlearning \cite{ha2019reinforcement}. By exploring randomly, an agent can identify what actions and states need to be unlearned or have their influence reduced.

Decremental methods can also effectively adjust to changes by learning to prioritize new information that reflects the current state of the environment\cite{li2022applications}. This makes them particularly useful in scenarios that are subject to frequent changes, such as adaptive control systems and real-time decision-making\cite{reddy2022reinforcement}. Overall, this ability to focus on recent data can lead to more effective learning and better performance in ever-evolving contexts. The incremental nature of DRL ensures that the forgetting process is controlled and does not adversely affect the agent’s performance in other environments. This is useful for maintaining the overall functionality of the model while ensuring privacy.

Another approach to RU is Environment Poisoning in which the unlearning environment itself is modified \cite{rakhsha2021policy}. This method aims to influence the agent’s policy of learning by creating a situation where its previously learned knowledge becomes less effective\cite{xu2021transferable}. The first step is to introduce changes for how the environment responds to the agent’s actions, effectively modifying the state transitions. The second step determines how the agent adapts to the new transition dynamics and finds an optimal strategy to maximize rewards under the new conditions. The last step involves analyzing the agent’s behavior and making further adjustments to the transition function to continue the unlearning process.
This approach helps to maintain the agent’s overall performance and stability even as it unlearns in a certain manner. All of these steps are iterative.


The other most effective approach in RU is Quantized Reward Konditioning (Quark), an algorithm designed to optimize a reward function that quantifies irrelevant data while maintaining the integrity of the original model. Quark is a RL technique that aims to unlearn certain properties from language models through three iterative stages: (1) exploration (sampling text with the current model, evaluating its reward, and storing it in a data pool), (2) quantization (sorting the data pool by reward and dividing it into quantiles), and (3) learning (updating the language model using samples from each quantile).

Quark helps refine the reward structure of a RL model, enabling it to discard or adjust behaviors that are no longer desirable without deviating too far from the original model’s function. Because of this attribute, Quark is a practical and efficient method for improving the performance and accuracy of large language models (LLMs), particularly when there is a need to remove biases or errors learned during pretraining.

\subsection{Continual Learning via Selective forgetting}
\label{sec:Continual and Selective} 
Continual Learning focuses on adapting to new knowledge while maintaining previous information\cite{chatterjee2024unified}. This can be done more accurately using Selective Forgetting. Selective Forgetting allows the system to unlearn specific tasks requested by a user without affecting the overall model performance \cite{brodzinski2024survey}. This saves time and space after every upgrade of a model. The major objective is to continue learning using Selective Forgetting without using the original data from previous tasks\cite{wang2024selective}. Selective Forgetting is indeed a significant challenge in LLMs due to their complexity and vast knowledge\cite{gundavarapu2024machine}.

Introducing Selective Pruning for LLMs is a promising approach for resolving this challenge \cite{pochinkov2024dissecting, shibata2021learning}. This method selectively removes abilities from trained LLMs \cite{qu2024frontier} by performing structured pruning.
The approach involves iteratively pruning nodes in either the feed-forward layers or attention head layers of the Transformer architecture. A scoring function evaluates the relative importance of nodes with respect to two datasets;
Forget Dataset (D-forget) and Retain Dataset (D-retain). D-forget is the dataset from which we want the model to forget capabilities\cite{li2023selective}.
D-retain is the dataset that the model should continue to perform well on. The major goal is to reduce the performance on D-forget while maintaining the performance on D-retain. This is a crucial step in the selective pruning method as it saves time while preserving the overall performance and structure of the model. It is a particularly valuable approach in real-world applications where models must be quickly adapted to new requirements without extensive retraining\cite{wang2024machine}. By implementing the ability to forget selective sensitive information, organizations can build trust with users through assuring them that their private data is handled carefully. 

The second method for unlearning is Decreasingly Matching the Moment (DMM), which treats the model parameters as a Gaussian distribution \cite{ziomek2024efficient}. DMM works by adjusting mean and variance of certain data points. This ensures that the impact of unlearning data points is minimized. The updates are calculated to match the moments (mean and variance) of the distribution before and after unlearning. All the steps in this process achieve unlearning by incrementally adjusting the model parameters \cite{zhang2023machine}.
By using a probabilistic approach, DMM can also incorporate privacy-preserving techniques, such as adding noise to the parameter updates, to ensure that the unlearning process does not leak sensitive information. This method is efficient because it avoids the need to retrain the entire model from scratch. Instead, it incrementally updates the parameters, making it faster and more practical for large-scale models\cite{heng2024selective}.
This method is particularly useful in scenarios where frequent updates or removal of data are required, and helps maintain the model’s performance while ensuring data privacy \cite{tarun2023deep}.

Selective Amnesia is also a contributory technique in Continual Learning, which helps the model to retain essential knowledge while forgetting less relevant or less important information. This technique maintains a balance between stability (preserving old knowledge) and adaptability (learning new knowledge) of the model\cite{heng2024selective}. The model may forget specific information of previous tasks that are no longer necessary for the performance of current tasks. This allows the model to avoid overfitting and ensures that it does not waste resources in retaining irrelevant information. Selective Amnesia involves dynamically identifying what parts of the network or memory are not important for the new task and thus can be forgotten without negatively impacting on the model's performance. Selective Amnesia is a way to design systems that can learn continuously and adapt to new tasks while avoiding the harmful effects of forgetting valuable information learned earlier.

\subsection{Corrective Unlearning}
\label{sec:Corrective}
Corrective Unlearning is a method designed to remove the influence of manipulated data from a trained model\cite{cha2024learning}. This approach is particularly useful when only a subset of the corrupted data can be identified and retrained. Applying this approach will increase the productivity of large-scale datasets\cite{goel2024corrective}.

Corrective Unlearning introduces Selective Synaptic Dampening (SSD) \cite{schoepf2024potion}, which can feasibly erase the influence of manipulated data when only a small representative of subset of data is identified. SSD is a method that relies on understanding how manipulated data influences the model's learned weights. This approach uses Fisher Information Matrix to assess what weights are most affected by the data \cite{schoepf2024parameter}. By dampening the influence of specific manipulated data points, SSD can enhance the robustness of models against attacks that exploit those weaknesses. This is an adaptive approach that allows for dynamic alterations in the model's learning process, making it more flexible for handling new information while managing the effects of older data \cite{foster2024loss}. The ability to effectively remove manipulated data has powerful implications for privacy, especially in contexts where models may have accidentally learned from biased or harmful data.

The second method that is gaining attention for its efficiency in corrective unlearning is Cognac. Cognac is a graph unlearning method that focuses on efficiently removing the influence of specific data points, known as the manipulation set, from Graph-based Neural Networks (GNN) \cite{kolipaka2024cognac}. Cognac works by identifying a small portion (e.g. 5\%) of a manipulation set that includes datasets that need to be unlearned. It visualizes the graph structure and investigates connections and relationships within the model. Instead of retraining a model from scratch, Cognac intelligently adjusts the influence of a manipulation set. This is done through targeted updates that minimize the impact of the identified data points on the model’s performance\cite{yang2023contrastive}.

The potential applications of Cognac are broad, including enhancement of data privacy, improved model robustness, and ensuring compliance with data regulations like GDPR\cite{wu2023graphguard}. It is particularly useful in scenarios where data points need to be removed due to errors, biases, or privacy concerns.

The third technique is Corrective UnRanking Distillation (CuRD)\cite{hou2024neural}. CuRD uses a novel teacher-student framework specifically designed to address corrective unranking in neural Information Retrieval (IR). CuRD enables a neural IR model to selectively forget specific samples (such as documents or relevant information) while ensuring that the ranking output for other samples remains unchanged. Its primary objective is to adjust the relevance scores of the documents to be forgotten so that their relevance becomes indistinguishable from low-ranking or non-retrievable documents, effectively rendering the forgotten documents invisible or irrelevant to the retrieval system.

Furthermore, CuRD ensures that the model continues to perform well on the samples not targeted for forgetting, preserving the overall integrity of the retrieval system. Instead of simply removing forgotten documents, substitute documents are introduced to take their place. These substitutes are designed to have relevance scores that closely match those of the original (forgotten) documents. This substitution process is crucial, as removing a document without replacing it could disrupt the ranking system’s integrity. The relevance scores of these substitute documents are fine-tuned during training to align with the relevance scores of the to-be-forgotten documents, ensuring that the model maintains a consistent and coherent ranking post-forgetting. This framework can be applied to various retrieval systems based on neural networks, such as search engines or recommendation systems, enabling compliance with unlearning requirements (e.g. privacy concerns) without compromising overall system performance.

\subsection{Federated Unlearning}
\label{sec:Federated}
Federated Learning (FL) is a decentralized approach to ML where models are trained across multiple devices (or clients) without sharing their private data\cite{banabilah2022federated},\cite{gao2024verifi}. In FL, each device uses local data for local training, then uploads the model to the server for aggregation, and finally the server sends the model update to the participants to achieve the overall learning goal\cite{wen2023survey}. FL \cite{romandini2024federated}appears to be a promising ML technique for keeping local data private \cite{mammen2021federated}. Federated Unlearning (FU) is an essential development in FL that addresses privacy and data management challenges\cite{halimi2022federated}. The primary objective of FU is to delete specific data requested by a client from a model that was trained using FL techniques\cite{dhasade2023quickdrop}.

To unlearn in FL is difficult because it requires retraining the model without a specific client's data, or clients that need to be forgotten \cite{wang2023federated,varshney2025efficient}. This can be computationally expensive and time-consuming\cite{li2020preserving}. Thus, FedEraser addresses this problem by utilizing the historical updates of model parameters from the federated clients\cite{zhao2023survey}. FedEraser stores the parameter updates (gradients or model changes) sent by each federated client during the training process\cite{liu2020federated}. In this method, the central server does not require all raw data to be stored, but only gets updates that are generated by clients during training of a model\cite{varshney2024efficient}. FedEraser does not use any information of target
clients, thus preserving privacy issues in unlearning\cite{liu2021federaser}.
FedEraser is designed to eliminate clients data on a global learning model while minimizing the time to construct an unlearning model. This technique is helpful in circumstances where unlearning is frequent or there is poisonous data. 

Another comprehensive unlearning technique in FL is Federated Adaptive Unlearning (FedAU) \cite{gu2024unlearning}. FedAU works by integrating a lightweight auxiliary unlearning module into the model. This module helps to efficiently remove the influence of specific data points or clients from the model, and facilitates unlearning without the need for extensive retraining\cite{xiang2024empowering}. FedAU helps multiple clients implement unlearning concurrently. It is an effective model accuracy technique that shows strong performance in unlearning.

FedAU shows promising achievements in FL, particularly in privacy preservation and model integrity\cite{rodio2024fedstale}. It works on a selective adaptation that ensures models can dynamically adapt to unlearning requests without the computational cost of retraining\cite{li2023federated}.Thus, it makes a model highly effective and provides a scalable solution for data privacy concerns in distributed ML environments\cite{wang2023lightweight}.

Quantized Federated learning (Q-FL) \cite{xiong2023exact} is considered one of the most effective algorithms for implementing exact FU while ensuring model convergence. Q-FL utilizes the exact FU\cite{tao2024communication} technique to enable efficient data deletion. With this approach, each device locally trains a model on its private data. After a local update, the model parameters are quantized and sent to a central server for aggregation. The quantization step helps reduce communication overhead by lowering the precision of the model updates without significantly affecting the model’s performance.
In FL, tracking the influence of specific data points on the global model is challenging. For exact unlearning to work, the system needs to monitor how each data point contributes to the model’s parameters. This can be achieved using gradient tracking mechanisms, which involve tracking local model updates linked to specific data points. This may involve associating the updates with the corresponding data points on each device. Once the unlearning process is completed, the model parameters are updated and re-quantized to maintain communication efficiency. The updated quantized model is then shared with all participating devices for further training or use.

\section{Advanced Metrics for Evaluating Unlearning and Datasets}
\label{evaluation and datasets}
\subsection{Evaluation Metrics}\label{sec:evaluation}

\begin{table*}[b]
\centering
\caption{Table of Evaluation Metrices}
\label{tab:evaluation metrices}
\begin{tabular}{|l|l|p{9cm}|l|}
\hline
Class                        & Evaluation Metrics        & Description                                                                                & Ref. \\ \hline
\multirow{3}{*}{Time}        & Running Time              & It is used to assess the efficiency of an algorithm.                                      &   \cite{liu2024model}   \\ \cline{2-4} 
 &
  Relearn time &
  It refers to the amount of time required for a model to re-acquire knowledge after forgetting certain data. & \cite{shaik2024exploring}
   \\ \cline{2-4} 
                             & Forgetting Rate           & Speed in which a model loses previously learned information. 
     $ Frate = \frac{\Delta L}{\Delta t} \label{eq:frate} $                               
    &  \cite{wang2024comprehensive},\cite{ma2022learn}    \\ \hline
\multirow{3}{*}{Performance} & Accuracy                  & The proportion of correctly classified instances in the test set. \hfill
   $ \text{Accuracy}_{\text{unlearned}} = \frac{\sum_{i=1}^{N} \mathbf{1}(y_i = \hat{y}_i)}{N}
    \label{eq:accuracy_unlearned} $       
                            & \cite{golatkar2020eternal,shaik2024exploring}    \\ \cline{2-4}
                             & Completeness              & The goal is to minimize the forgetfulness loss (means that forgotten data is fully unlearned) and maintain a high accuracy on the remaining data. $
    \text{Completeness} = \lambda_1 \times \text{Loss \ of \ forget \ data} + \lambda_2 \times (1 - \text{Accuracy}_{\text{retain}})
    \label{eq:completeness}$                                   &   \cite{sai2024machine}   \\ \cline{2-4} 
                             & Epistemic Uncertainty     & Refers to uncertainty due to lack of information about a model, which can be reduced. 
                             $ \text{Efficacy}(w, D) = 
    \begin{cases} 
        \frac{1}{i(w; D)}, & \text{if } i(w; D) > 0 \\
        \infty, & \text{otherwise} 
    \end{cases}
    \label{eq:efficacy}$      & \cite{nguyen2022survey}     \\ \hline
\multirow{2}{*}{Resources}   & Storage Requirement       & Amount of space needed to store data or resources 
    $ S_{\text{unlearning}} = S_{\text{model}}^{\text{before}} + S_{\text{data\_removed}} + S_{\text{intermediate}} + S_{\text{model}}^{\text{after}}
    \label{eq:unlearning} $                                         &  \cite{xu2024machine}    \\ \cline{2-4} 
                             & Incorporation Cost        & The costs incurred while updating the concept descriptions with a single training instance. $
    C_{\text{incorporation}} = C_{\text{data}} + C_{\text{computational}} + C_{\text{model update}} + C_{\text{validation}} \label{eq:incorporation}$ & \cite{xu2024machine}     \\ \hline
\multirow{2}{*}{Verification } &
  Theory-based Verification &
  Formal mathematical concept and logical reasoning to prove accuracy of a algorithm & \cite{tu2024towards}
   \\ \cline{2-4} 
                             & Attack-based Verification & It tests a system's security and asseses its ability to resist threats.                        &   \cite{wang2024comprehensive}   \\ \hline
\end{tabular}
\end{table*}

There are several metrics used to ensure that the unlearning process is effective and efficient, thus maintaining the model's integrity and performance. In our survey, we suggest the following metrics to verify unlearning. These matrics are also summarized in th table \ref{tab:evaluation metrices}

\subsubsection{Accuracy}
\label{sec:accuracy}
The unlearning process should remove all traces of irrelevant data, ensuring that the model no longer memorizes or uses this data when making predictions\cite{liu2024machine}. In IU, a model’s accuracy can be compared against three different datasets: 1) Forget set: ensures that the model has forgotten the specific data that is no longer needed. 2) Retained Set: ensures that the model has retained all the  useful information after unlearning has done. 3) Test Set: ensures that the unlearning process does not negatively affect the model's generalization\cite{huang2024learning}.

\subsubsection{Running Time}
\label{sec:running}
The timeliness of unlearning has a significant advantage over retraining, especially when there is a need to quickly restore privacy and security in a model. Unlearning enables faster model updates because it requires fewer computations by targeting only the forgotten data or a limited set of model parameters, compared to retraining using the full dataset\cite{tarun2023fast}\cite{sai2024machine}. The speed-up factor of unlearning over retraining can be substantial, particularly when the unlearning operation focuses only on small, targeted updates to the model rather than needing to process all the data from scratch.

\subsubsection{Theory-Based Verification} \label{sec:theory}
Theory-based verification in IU is crucial in ensuring that the unlearning process maintains model integrity and preserves model performance and privacy. By applying formal methods like differential privacy, influence functions and model checking, the unlearning process can be verified\cite{li2024machine}\cite{tu2024towards}.
The major objective of IU is to ensure that a model no longer uses the data points that were not learned. Theory-based verification in this context should ensure that once data points are removed they can no longer influence the model's predictions.

\subsubsection{Attack-Based Verification}
\label{attack}
This ensures that the model cannot leak any information about the unlearned data. If the model remains robust against these attacks, it means that unlearning was successful in erasing the influence of the data while preventing leakage of sensitive information.
Attack-based verification of IU is essential to ensuring that the unlearning process is resilient against adversarial attacks\cite{wang2024comprehensive}\cite{tu2024towards}. By simulating various types of attacks, such as data reconstruction, model inversion, and adversarial input manipulation, attack-based verification can help identify vulnerabilities and provide insights into the robustness of unlearning methods. This helps ensure that the unlearning process effectively removes the influence of specific data points, preserves privacy, and prevents malicious exploitation.

\subsubsection{Forgetting Rate}
\label{sec:forgetting}
To evaluate the forgetting rate in IU, the general focus is how effectively the model forgets the data it is supposed to forget, how much its performance on retained data changes, and how it generalizes on new, unseen data. Metrics such as forgetting rate provide quantitative measures of the success of the unlearning process\cite{wang2024comprehensive}. Additionally, it is crucial to track computational efficiency to ensure that the unlearning process is both effective and functional.

\subsubsection{Storage Requirement}
\label{storage}
The storage requirements in IU depend on several factors, which can vary based on things like unlearning methods, the size of the model, and the data involved. In an unlearning situation, the model parameters may need to be modified to reflect the removal of specific data points. This may require additional storage space to store updated versions of the model. IU involves saving different versions of the model (e.g., pre- and post-unlearning), which requires additional storage to store these changes. The major aim of efficient unlearning techniques is to minimize storage by modifying only relevant parts of the model using compressed data representations, minimizing the need for full retraining.

\subsubsection{Completeness}
\label{sec:completeness}
This is a critical aspect that measures how effectively the unlearning process has removed the influence of forgotten data and ensures that the unlearned model behaves in a similar fashion to a retrained model. High completeness of the rate indicates that the unlearned model has effectively forgotten the erased data\cite{sai2024machine}.  Jaccard distance is used to calculate the overlap of the output space between the unlearned and retrained models. The lower the Jaccard distance, the higher the completeness.

\subsubsection{Epistemic Uncertainty}
\label{sec:epistemic}
Epistemic Uncertainty refers to the uncertainty about the model's parameters or structure due to a lack of knowledge or data. This can be thought of as model uncertainty\cite{wang2024comprehensive},\cite{sai2024machine}. In IU, the model is continuously updated to forget certain data points while retaining its generalization capability on the remaining data. This process inherently introduces uncertainty into the model's behavior, and assessing this uncertainty is crucial for ensuring that unlearning has not adversely affected the model's performance, especially when it is supposed to "forget" specific data but still perform well on other data.

\subsubsection{Incorporation Cost}
\label{sec:incorporation}
The Incorporation Cost in IU is a measure of the time, resources, and parameter adjustments required to update a model as it forgets certain data. By investigating time complexity, computational resources, and the extent of parameter adjustments, it can be estimated how much unlearning process is costing\cite{zhao2023survey}. Optimizing unlearning algorithms can reduce the costs by focusing on partial updates, fine-tuning, and efficient data processing. This can make the unlearning process much faster and less resource-intensive, which is crucial in real-world applications requiring frequent updates and facing privacy issues.

\subsubsection{Relearn Time}
\label{sec:relearn}
This metric will help to assess how much information the model retains after the unlearning process, particularly when a user needs to verify that sensitive or unwanted data has been properly removed\cite{tarun2023fast}, \cite{sai2024machine}. It provides a significant method for checking if the unlearning process has been completed by measuring how quickly the model recovers information about the unlearned data.

\subsection{Datasets}

We compile the commonly used datasets in MU research and provide a detailed introduction to each of them in Table \ref{tab:dataset}. These datasets generally fall under four categories: Image, Tabular, Text, and Graph. Image datasets are the most widely used in unlearning studies, typically used to train classification models. On the other hand, tabular datasets are primarily applied in recommendation systems. Unlearning research focused on recommendation models often leverages these tabular datasets. Graph data is used for tasks like node classification and link prediction, and it is commonly featured in graph unlearning studies.

\begin{table*}[!h]
\centering
\caption{List of Datasets}
\label{tab:dataset}
\begin{tabular}{|c|c|c|c|c|}
\hline
\textbf{Dataset} & \textbf{Modality}                & \textbf{No. of Instances} & \textbf{No. of Attributes} & \textbf{Task}     \\ \hline
MINIST\cite{zhang2023machine}                     & \multirow{5}{*}{Image}   & 70,000      & 784    & Object Recognition      \\ \cline{1-1} \cline{3-5} 
CIFAR-100 \cite{goel2024corrective}\cite{zuo2024ecil}                 &                          & 60,000      & 3072   & Object Recognition      \\ \cline{1-1} \cline{3-5} 
SVHN \cite{hu2024eraser}                      &                          & 600,000     & 3072   & Object Recognition      \\ \cline{1-1} \cline{3-5} 
ImageNet  \cite{cha2024learning}                 &                          & 1.2M & 1000   & Object Recognition      \\ \cline{1-1} \cline{3-5} 
Mini-ImageNet \cite{hoang2024learn}             &                          & 100,000     & 784    & Object Recognition      \\ \hline
MovieLens \cite{vacchetti2022movie}                 & Recommendation           & 100,000     & varies & Movie Recommendation    \\ \hline
Human Activity Recognition\cite{islam2022human} & Sequence                 & 10,299      & 561    & Activity Recognition    \\ \hline
Transaction  \cite{su2023asynchronous}               & Purchase                & 39,624      & 8      & Purchase Prediction     \\ \hline
Fashion-Minist \cite{kim2024layer}   & \multirow{2}{*}{Computer Vision} & 70,000                    & 784                        & Image Recognition \\ \cline{1-1} \cline{3-5} 
Youtube Faces \cite{bonettini2021video}             &                          & 3,425       & 2622   & Face Recognition        \\ \hline
OGB  \cite{hu2020open}                      & \multirow{2}{*}{Graph}   & 1.2 M       & varies & Graph Classification    \\ \cline{1-1} \cline{3-5} 
Cora  \cite{chien2022efficient}                     &                          & 2,708       & 1433   & Graph Classification    \\ \hline
Activity Recognition \cite{fan2023machine}      & Time Series              & 10,299      & 561    & Activity Classification \\ \hline
Adult \cite{chen2024fast}                     & \multirow{3}{*}{Tabular} & 48,842      & 14     & Income Prediction       \\ \cline{1-1} \cline{3-5} 
Breast Cancer\cite{michalski1986multi}              &                          & 286         & 9      & Cancer Diagnosis        \\ \cline{1-1} \cline{3-5} 
Diabetes \cite{warnecke2021machine}                  &                          & 768         & 8      & Diabetes Diagnosis      \\ \hline
\end{tabular}
\end{table*}

\section{Challenges Of Incremental Unlearning}\label{challenges}
ML models can accidentally memorize sensitive, unauthorized, or malicious data during training, which creates potential risks in terms of privacy, security vulnerabilities, and performance degradation \cite{xu2024machine},\cite{li2024machine}. These issues are particularly observed in scenarios where data privacy is a concern, such as healthcare, finance, and personalized services\cite{liu2021machine},\cite{li2019impact}.

\begin{figure}[htb]
    \centering
    \includegraphics[width=\columnwidth]{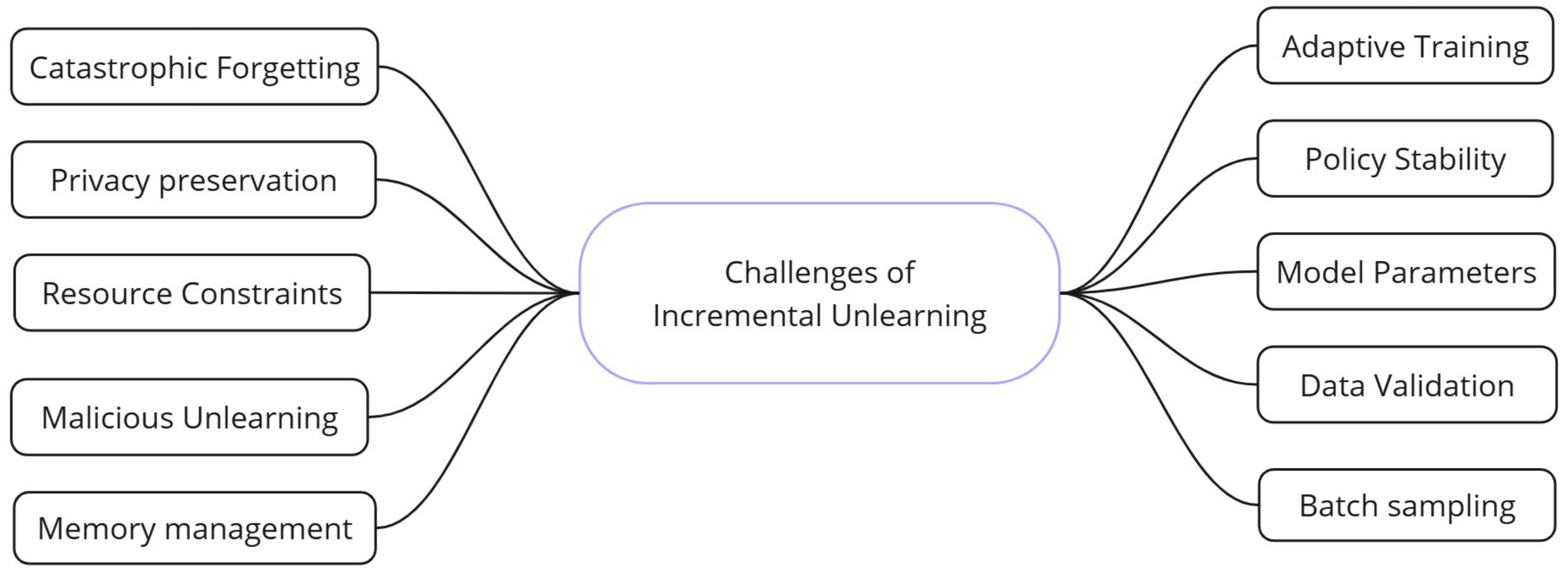}
    \caption{Incremental Unlearning Challenges}
    \label{figs: Inc UL challenges 2}
\end{figure}

Here, we highlight some of the main challenges of IU. These challenges are also summarized in figure \ref{figs: Inc UL challenges 2}.

\subsubsection{Adaptive Training} \label{sec:adaptive training}

When we want to remove a specific data sample, the challenge is to erase its influence on the model without affecting the model's behavior on other data points\cite{gupta2021adaptive}. This is not an easy task because once a sample has contributed to the model’s parameters, removing that sample requires altering the entire learning path\cite{guo2019certified}. In many MU models, especially DNNs, the relationship between a data sample and the model’s parameters is non-linear. This means that the effect of removing a single data point is not just a simple reverse of an update, but a complex adjustment to the model’s state\cite{reddy2024comparison}.

\begin{itemize}
\item \emph{Adaptive Adjustments of Data:}\\
When we remove certain data points, the distribution of the remaining data may change. The model must adapt to these changes, which can modify the relationships between data features and predictions. This can lead to the challenge of data shift where the model must adjust to a new data distribution without retraining entirely.
Different strategies for unlearning have varying effects on how well the model adapts. Some strategies may cause the model to lose important patterns that are not linked to the removed data, which impacts the model’s overall adaptive capacity\cite{gao2024practical}.

\item \emph{Maintaining Model's Performance:}\\
After removing specific data, the model must continue to perform well on the remaining data\cite{shaik2024exploring}. The challenge lies in ensuring that the model forgets the influence of the unlearned data without forgetting useful knowledge. If removal is not handled properly, unlearning can lead to catastrophic forgetting, where the model’s generalization ability on other data will diminished. Unlearning must not affect the model’s ability to generalize to new, unseen data. 

\item \emph{Handling Multiple Iterations:}\\
In real-world scenarios, unlearning may need to be done incrementally over multiple steps as more data is removed. Each iteration of unlearning must simultaneously adapt the model appropriately without causing a degradation in performance or introducing biases\cite{shaik2024framu}. The challenge is to ensure that each incremental update maintains the integrity of the model’s overall performance over time.
\end{itemize}

\subsubsection{Memory Management}\label{sec:memory management}

Another issue is the lack of memory of historical updates, which poses a significant challenge in unlearning\cite{sekhari2021remember}. Since the model updates its parameters incrementally with each new data point, it does not retain a history of how each parameter was influenced by specific data points. This makes it difficult to reverse the effects of a particular data sample.

\begin{itemize}
\item \emph{Avoid Excess Memory Usage:}\\
When data points are removed, there is a risk that some parts of the model will continue to either hold onto unnecessary data or contain references to removed data. This causes unnecessary memory expansion, where memory is allocated but not actively used. In some cases, memory management must ensure that unused or non-relevant references are cleared, so that space is recovered to avoid inefficient use of resources\cite{blanco2024digital}.

\item \emph{Handling Large Models:}\\ 
Large-scale models (e.g. DNNs with millions of parameters) are especially prone to memory management issues when performing IU. When large amounts of data are removed, portions of the model’s weights may need to be altered. This creates a challenge in updating only the necessary parts of the model without wasting memory resources\cite{xu2024machine}.
\end{itemize}

\subsubsection{Catastrophic Forgetting}\label{sec:catstrophic}

One of the biggest risks of unlearning specific data points, particularly in neural networks or other complex models, is catastrophic forgetting\cite{golatkar2020eternal,belouadah2021comprehensive}. This refers to the phenomenon where the model forgets not only the data that is being unlearned but also its relevant knowledge. This is particularly difficult when multiple data points are unlearned. 

\begin{itemize}

\item \emph{Task-based Forgetting:}\\ 
In some scenarios, unlearning may need to be task-specific, meaning the model must forget data connected to one specific task but retain its ability to perform on other tasks. For example, in a multi-task scenario, one task's data may need to be unlearned while ensuring that the model continues to perform effectively on other tasks\cite{izzo2021approximate}. Catastrophic forgetting in such settings occurs when the shared model parameters that affect multiple tasks are adjusted such that they reduce the model’s ability to perform perfectly on tasks not directly linked to the unlearned data.

\item \emph{Balancing Forgetting and Generalization:}\\ 
Unlearning techniques should ensure that the model remains generalizable to new data while removing the effects of the unlearned data. However, balancing generalization with forgetting is a difficult task. If the model forgets too much, it could become non-relevant to the remaining data, losing its ability to generalize effectively to new, unseen data. Conversely, if only a small amount of data is forgotten, the model may not meet the unlearning objectives, such as ensuring privacy or compliance with legal regulations\cite{schioppa2024model}. This balance is particularly difficult in the context of IU, where unlearning must be performed progressively over time, otherwise there will be performance degradation.
 
\item \emph{Efficient Unlearning:}\\ 
Unlearning should be computationally efficient to avoid the need for retraining the entire model from scratch. It is crucial to verify that the data has been completely and effectively unlearned. This consistency is essential to maintain trust in the predictions of the model\cite{chen2023unlearn}.
\end{itemize}

\subsubsection{Resource Constraints for Large models}\label{sec:resource mangement}

Unlearning specific data points can be computationally expensive, particularly for large models with millions of parameters\cite{zhou2024limitations}.The process of unlearning often requires additional memory to store intermediate states or checkpoints\cite{schwinn2024soft}. The training process in large models is highly interdependent, meaning that changes to one part of the model can affect other parts\cite{liu2024machine}. Unlearning should be resource-efficient to avoid the need for retraining the entire model from scratch.

\begin{itemize}
\item \emph{Time Constraints:}\\
IU can be time-consuming, particularly when dealing with large datasets or models. Unlike traditional training where updates are applied uniformly, unlearning involves precise modifications to the model that take time. The process must be efficient enough to comply with real-time requirements, especially when the unlearning involves urgent privacy requests\cite{tarun2023fast}.

\item \emph{Space Constraints:}\\
Certain scenarios of IU requires maintenance of additional data or versions of the model over time to ensure data has been properly forgotten \cite{yao2023large}. This adds to storage requirements, particularly in resource-limited environments such as mobile devices or edge computing devices where storage space is limited.

\item \emph{Cost Constraints:}\\
IU may involve modifying the model's parameters or architecture without retraining from scratch. This requires refined algorithms and significant computational resources to process updates efficiently\cite{si2023knowledge}.
\end{itemize}

\subsubsection{Privacy Preservation}\label{sec:privacy}

In Machine learning as a Service (MLaaS), the service provider often does not have direct access to the user's data\cite{liu2024threats}. This separation can complicate the unlearning process, as the provider must ensure that the data is effectively removed without directly handling it\cite{hu2023duty}. Ensuring that unlearning does not introduce or worsen biases in a model is essential. When the model developer and service provider are distinct entities, ensuring the privacy of data designated for unlearning becomes complex.

\begin{itemize}

\item \emph{Auditability and Transparency:}\\
Transparency in how data was deleted is essential for compliance with privacy regulations. Systems must provide clear auditing mechanisms to verify whether data has been properly removed and that the model no longer retains traces of sensitive information. This challenge becomes more complex in IU, as changes to the model may not be easy to trace or verify over time without full transparency about the unlearning process\cite{vidal2024verifying}.

\item \emph{Security Risk:}\\
Addressing potential security risks (such as adversarial attacks) that can exploit the unlearning process to infer deleted data is critical. This requires robust security measures to protect against such vulnerabilities\cite{marchant2022hard}.
\end{itemize}

\subsubsection{Malicious Unlearning}\label{sec:malicious}

Malicious unlearning occurs when an attacker submits requests to remove specific data from the model’s training dataset, with the intention of either degrading model's performance or embedding a backdoor into the model. Since the data targeted for removal may seem neutral or non-malicious, it becomes challenging to distinguish between valid unlearning requests and malicious ones\cite{hu2023duty}. The attacker could target specific data that the model heavily relies on, thus reducing its effectiveness without raising immediate alarms. Addressing these threats requires advanced techniques and continuous research to ensure the security and reliability of MU systems.

\begin{itemize}
\item \emph{Information Leakage:}\\
Malicious unlearning can lead to information leakage, where sensitive data that should have been forgotten is still accessible from the model\cite{he2021secure}. This poses significant privacy risks.

\item \emph{Data Injection:}\\ 
An attacker may intentionally introduce harmful data into the system intended to disrupt the unlearning process. This could result in the unlearning of incorrect data or the failure to unlearn sensitive data\cite{tang2023ensuring}.

\item \emph{Model Degradation:}\\
Malicious unlearning may attempt to cause model parameters to update in ways that introduce weaknesses, such as increasing susceptibility to certain types of attacks or degrading model performance\cite{chen2017targeted}.
\end{itemize}

\subsubsection{Model Parameters}\label{sec:model parameters}

One approach to unlearning is altering the model's parameters to remove certain data points. This requires careful adjustments to ensure that the model forgets the specific data without degrading its overall performance\cite{nguyen2022survey}. Maintaining the model's ability to generalize perfectly to new data is crucial. Unlearning specific data points or specific clients should not negatively impact the model's accuracy on the remaining data.

\begin{itemize}
\item \emph{Retaining While Forgetting:}\\ 
There are some unlearning methods that attempt to forget a specific part of the data while preserving the rest of the knowledge. However, this is difficult because the parameters are often shared across different data points. For example, a neural network may have learned some patterns that involve relationships between multiple data points, and removing one can have side effects on parameters that affect other data points\cite{liu2024towards}.

\item \emph{Complexity of Updates:}\\
There is no direct way to reverse the impact of a specific data point on the model parameters. Unlearning requires the model to adjust its weights or internal representations in a way that unties the influence of the data without retraining the entire model\cite{xu2024machine}.
If the model has been updated multiple times incrementally, the changes due to each update may be deeply intertwined, making it harder to isolate the effect of specific data points. Therefore, unlearning data without disturbing the learned model becomes more difficult as the model progresses.
\end{itemize}

\subsubsection{Data Validation}\label{sec:data validation}

Data Validation is a major concern as models become more complex and datasets expand in size\cite{gao2016big}. Data Validation in unlearning ensures that the data to be removed is correctly identified and that the removal process does not accidentally degrade the model’s performance or introduce malicious vulnerabilities\cite{fan2025challenging}. In large-scale models, the amount of data and the number of parameters involved makes it difficult to validate (a) whether the correct data has been removed and (b) whether the unlearning process has been successful without involving inefficiencies. Moreover, the increasing size of datasets means that traditional validation methods could become computationally expensive or infeasible.

\begin{itemize}
\item \emph{Ensure Accurate Data Removal:}\\
An important challenge in unlearning is to ensure that only the correct data is removed while maintaining the integrity of the model. It is critical to avoid mistakes in unlearning that could either over-remove data  or under-remove data\cite{paniv2024unsupervised}. Over-removing data can cause the model to lose important knowledge or generalization capability, while under-removing data allows some data to persist in the model when it should have been forgotten.

\item \emph{Dealing with Large Data Streams:}\\
In large-scale models, the amount of data and the number of parameters involved make it difficult to validate whether the correct data has been removed and whether the unlearning process has been successful without involving inefficiencies \cite{paniv2024unsupervised}. Furthermore, the increasing size of datasets means that standard validation methods could become computationally expensive or infeasible.

\item \emph{Consistency Between Data and Model:}\\
As a model learns incrementally, the data and the model's learned parameters changes over time. After unlearning, there must be consistency between the model and the data. Ensuring this consistency is particularly challenging, since the model’s learned knowledge can be influenced by a complex set of data points\cite{qu2023learn}. Unlearning one data point could have certain consequences on the model’s generalization.
Model drift might occur after data removal, which means that the model's behavior may shift in unpredictable ways if linked relationships within the data are disturbed.
The data used for unlearning may interact with the model’s learned features, therefore removing certain data points can break the link of reliable data patterns of the model.
\end{itemize}

\subsubsection{Batch Sampling}\label{sec:batch sampling}

Effective batch sampling is critical for handling large datasets, as failures in the sampling process can lead to increased computational costs and slower unlearning\cite{kurmanji2024towards}. If sampling is inefficient, the process of updating the model to reflect the unlearning could become remarkably slow or resource-intensive. Batch sampling methods need to identify subsets of data to modify or remove in a way that minimizes the overall computational concern. When it comes to unlearning in adaptive models, this may cause serious difficulties.

\begin{itemize}
\item \emph{Handling Dependencies:}\\
In Reinforcement Learning, unlearning data in a batch might involve forgetting certain episodes or state-action pairs. Since Reinforcement Learning depends heavily on the cumulative history of actions and rewards, removing a batch of episodes could affect the policy in unexpected ways and disable the agent’s learned behavior\cite{dulac2021challenges}.
In time-series applications, data from earlier points in the series can directly influence later data. Removing a batch of data from an earlier time period may cause significant shifts in the model’s understanding of patterns, leading to potential inconsistencies when the model predicts future values.

\item \emph{Consistency across Batches:}\\
When unlearning is performed incrementally over time, ensuring consistency between batches can be difficult\cite{sai2024machine}. Each unlearning step may result in slight variations in the model's internal parameters, which can accumulate and lead to policy instability over time.
\end{itemize}

\subsubsection{Policy Stability}\label{sec:policy stability}
Removing knowledge of certain environments must be done in a way that it does not adversely affect the agent’s learning and decision-making abilities in other environments\cite{ye2023reinforcement}. When the agent forgets specific environments (or certain states, actions, or rewards), it can break the reliability of the learned policy. The agent might rely on certain features, states, or transitions that were only relevant in the unlearned environments, and their removal could disrupt its policy in other environments.

\begin{itemize}
\item \emph{Dynamic Environments:}\\
IU aims to remove the influence of specific data points, but this must be done while adapting to the changes in the environment. The challenge arises when these updates interfere with the agent's ability to respond effectively to progressing environmental conditions\cite{jaman2024machine}. For example, if the agent is required to forget certain experiences (such as specific traffic data), it must adjust its behavior while ensuring it can still adapt to new traffic conditions. Unlearning outdated data could disable its ability to navigate efficiently in new or changing conditions.

\item \emph{Delayed Adaptation:}\\
When unlearning is applied in incremental systems, the agent may need to relearn or adapt to the new environment, which takes time. The challenge is that the agent may not be able to convert to the new optimal policy after unlearning, leading to temporary performance dips or errors. For example, a climate prediction model may need to unlearn irrelevant weather data due to a policy change or data privacy requirements\cite{dulac2021challenges}. The unlearning process can disable the model temporarily as it struggles to adapt to the new conditions, affecting the predictions until it can completely adapt to the new changes.
\end{itemize}

\section{Potential Solutions and Future Directions}\label{sec:potential and future}
In this section we present the potential solutions and the future directions in the domain of IU.

\subsection{Potential Solutions}\label{subsec:potential solutions}

IU presents several challenges, which can be addressed to varying degrees by a variety of solutions. Various innovations can address the challenges of IU while ensuring that models remain adaptable, efficient, and compliant with data regulatory requirements. In this section, we analyze the current and potential developments in IU.
\begin{itemize}
\item \emph{Fine-Tuning:}\\  
An efficient strategy for IU is using a fine-tuning technique rather retraining the entire model\cite{han2024parameter,ding2024bridging,blanco2025digital}. By adjusting the model's parameters slightly, IU can remove specific data points without affecting the overall performance of the model\cite{zheng2024llamafactory}. A fine-tuned model demonstrates higher performance on all tasks along with enhanced generalization capabilities \cite{xu2021raise}.
    
\item \emph{Approximation Techniques:}\\ 
The approximation technique in machine unlearning revolves around finding efficient and accurate ways to remove specific knowledge from a model without requiring complete retraining. Using approximation techniques like stochastic gradient descent (SGD) \cite{chien2024stochastic} or gradient-based methods can make unlearning more efficient, as opposed to retraining the entire model from scratch \cite{lev2024faster}. By approximating the effect of data removal via gradients, the model can adjust its parameters slightly to forget the influence of specific data points without full retraining. These methods give priority to small and incremental updates to the model weights rather than a complete retraining process \cite{liu2024model,xia2024edge}.
    
\item \emph{Influence Functions:}\\
Influence Functions can help determine the influence of a specific data point on the model's predictions. By evaluating the impact of a data point, we can calculate its influence on the model \cite{zhang2023recommendation,chen2024fast}. This approach has been proposed as a way to remove data points while maintaining overall model performance\cite{liu2024model}.
    
\item \emph{Parallel and Distributed Unlearning:}\\
For large-scale datasets, IU can be scaled using parallel or distributed unlearning techniques \cite{hu2023eraser}. By distributing the unlearning task across multiple nodes, the time required to forget certain data points or concepts can be significantly reduced
\cite{liu2024fishers}.
    
\item \emph{Knowledge Distillation:}\\
This is a technique where a large, complex model is compressed into a simpler, more interpretable model \cite{kim2022efficient}. After unlearning, the knowledge distillation process can help preserve important model behavior and make the model easier to interpret post-unlearning \cite{wu2022federated,zheng2023graph}.
    
\item \emph{Data Partitioning:}\\
This can be a valuable strategy in the context of IU, particularly in improving efficiency and removing specific data points\cite{zeng2023towards}. If the data is well-partitioned, it may be possible to break the model's reliance on specific partitions. This modular approach makes it easier to remove certain data points or even entire partitions without affecting the rest of the model \cite{chen2022recommendation}. 
\end{itemize}

\subsection{Future Directions}
As data-controlled systems become increasingly widespread, the ability to efficiently, securely, and transparently forget data will become a crucial aspect of ML systems. By addressing the key areas outlined below, future research can greatly improve the practical implementation of IU, ensuring that AI systems comply with privacy regulations while preserving model performance, stability, and fairness.

\subsubsection{Natural Language Processing (NLP)}\label{sec:future direction nlp}

 Research on IU has primarily focused on specific application domains, such as healthcare and finance. However, its potential applicability in other areas, like natural language processing (NLP), speech recognition, or autonomous systems, has yet to be fully explored. One of the key objectives of ML models in NLP is to accurately predict or classify text data by identifying patterns and relationships from a training dataset. However, if the training data is incomplete, biased, or outdated, the model may learn incorrect or irrelevant patterns, leading to poor performance when applied to new data\cite{shaik2024exploring}. Further research is needed to explore how IU techniques can be generalized across various ML tasks and domains in NLP. A promising direction for future work might involve developing flexible frameworks that can be applied to a wide range of use cases, particularly those involving complex and multi-dimensional datasets\cite{shaik2022review}.

\subsubsection{Image Processing}\label{future direction ip}
In image classification, a model could learn to forget images from old categories while continually learning to classify new ones\cite{zhang2022machine}. This can be done incrementally by advancing ML frameworks such as Continual Learning, FL, and Transfer Learning. The innovations in efficiency, privacy, and real-time adaptation will make IU an essential component of next-generation image processing models, enabling them to maintain high performance while adhering to privacy regulations and ethical standards.
 
\subsubsection{Recommendation System (RS)}\label{subsec:rs}
 One area where IU can play a crucial role in the future of recommendation systems is in the development of more adaptive and personalized models. As the volume and complexity of user data increase, MU models will need to adapt to evolving user preferences and needs\cite{huynh2025certified}. IU can help these models stay up-to-date with accurate recommendations by enabling them to modify or eliminate irrelevant or misleading features \cite{ramscar2023discriminative}\cite{asatiani2021sociotechnical}. Additionally, addressing catastrophic forgetting is essential to ensure that unlearning specific data does not negatively affect recommendation accuracy\cite{shaik2024exploring}. 
 
 Research in this area should also aim to address issues of bias and fairness by allowing the system to unlearn biased data, leading to more equitable recommendations. There is a need to use unlearning strategies that continuously monitor the model and the data for emerging bias. Even as IU methods need to update a model as new data arrives, they must also be combined with periodic bias audits to ensure that any new biased features are removed as the data emerges. This approach can help to prevent producing bias into the model as it is adapting to new data. Future research on IU should help to protect user privacy in a recommendation system. By removing or identifying sensitive information from the data used to train these models, MU can ensure that a user’s personal information is not used inappropriately or disclosed without their consent.


\section{Conclusion}\label{conclusion}
This survey of research literature has taken significant steps to address the challenges in IU, particularly around the efficient and effective removal of specific data points from ML models without retraining from scratch. This area of research is motivated by the increasing importance of data privacy regulatory compliance (e.g. GDPR) and the need for models to adapt to changing environments or outdated data. IU techniques focus on minimizing computational overhead while ensuring that the model continues both to perform well and recognize privacy constraints.

One of the key contributions of IU research is the development of methods that allow models to unlearn data without compromising performance. Reviewed techniques such as decrement RL method, environment poisoning method, Quark, selective pruning method, selective amnesia, cognac, CuRD, FedAU, Q-FL and FedEraser have shown significant results in removing the influence of selected data points while retaining the model’s integrity. These techniques are useful for LLMs and for dynamic environments.

In conclusion, IU research is progressing rapidly, with practical applications in privacy-sensitive domains, as well as dynamic ML and unlearning environments. Although there are certain challenges like privacy, model efficiency, and robustness issues in unlearning, current advancements continue to make IU a powerful tool for responsible and adaptive AI systems.


{
\bibliographystyle{ieeetr}
\bibliography{S_Refs}
}

\end{document}